\documentclass[Afour,sageh,times]{sagej}

\usepackage{moreverb,url}

\usepackage[colorlinks,bookmarksopen,bookmarksnumbered,citecolor=red,urlcolor=red]{hyperref}

\usepackage{booktabs,tabularx,soul,color,framed}
\sethlcolor{yellow}
\soulregister\cite7
\soulregister\citet7
\soulregister\citep7
\soulregister\ref7
\soulregister\eqref7
\definecolor{shadecolor}{rgb}{1.0,0.94,0}
\newcommand\BibTeX{{\rmfamily B\kern-.05em \textsc{i\kern-.025em b}\kern-.08em
T\kern-.1667em\lower.7ex\hbox{E}\kern-.125emX}}

\def\volumeyear{2025}

\begin{document}

\runninghead{Hu et al.}

\title{Learning High-Fidelity Robot Self-Model with Articulated 3D Gaussian Splatting}

\author{Kejun Hu, Peng Yu and Ning Tan}

\affiliation{School of Computer Science and Engineering, Sun Yat-sen University, Guangzhou, Guangdong, China}

\corrauth{Ning Tan, School of Computer Science and Engineering,
Sun Yat-sen University,
Guangzhou,
Guangdong,
510006, China.}

\email{tann5@mail.sysu.edu.cn}

\begin{abstract}
Self-modeling enables robots to learn task-agnostic models of their morphology and kinematics based on data that can be automatically collected, with minimal human intervention and prior information, thereby enhancing machine intelligence. Recent research has highlighted the potential of data-driven technology in modeling the morphology and kinematics of robots. However, existing self-modeling methods suffer from either low modeling quality or excessive data acquisition costs in {equipment}. Beyond morphology and kinematics, surface color is also a crucial component of robots, which is challenging to model and remains unexplored. In this work, a high-quality, surface color-aware, and link-level method is proposed for robot self-modeling. We utilize three-dimensional (3D) Gaussians to represent the static morphology and surface color of robots, and cluster the 3D Gaussians to construct neural ellipsoid bones, whose deformations are controlled by the transformation matrices generated by a kinematic neural network. The 3D Gaussians and kinematic neural network are trained using data pairs composed of joint angles, camera parameters, and multi-view images without depth information. By feeding the kinematic neural network with joint angles, we can utilize the well-trained model to describe the corresponding morphology, kinematics, and surface color of robots at the link level, and render robot images from different perspectives with the aid of 3D Gaussian splatting. Furthermore, we demonstrate that the established model can be exploited to perform downstream tasks such as motion planning and inverse kinematics.
\end{abstract}


\maketitle

\section{Introduction}
High-quality models of geometry and kinematics are essential for effective robot planning, control, and visualization. Traditional modeling approaches demand substantial human effort to gather the necessary parameter information of the robot, and over time, the quality of these models tends to deteriorate due to various factors, such as unexpected damage, joint malfunctions, and wear. Therefore, the development of methods that enable robots to collect data and learn their own models autonomously is critical for ensuring the successful execution of tasks and advancing the level of machine intelligence. Inspired by human behavior \citep{rochat2003five}, the concept of robot self-modeling has recently gained attention among researchers \citep{chen2022fully,diaz2023machine}. A robot self-model is a general-purpose representation of a robot's physical shape and structure that can be acquired at any time without human intervention~\citep{kwiatkowski2022deep,schulze2024high}. Through self-modeling, robots can autonomously and continuously update their knowledge of both morphology and kinematics, which not only ensures that the model information remains up-to-date and adaptable to various unforeseen circumstances but also simplifies the modeling process for newly designed robots.

{Early approaches focused on designing task-specific self-modeling techniques,~\citep{chen2021smile,hang2021manipulation}
}such as modeling end-effector positions \citep{mathew2014learning}, tilt angles \citep{bongard2006resilient}, or joint velocities~\citep{sanchez2018graph} according to task requirements. However, due to the lack of generality and adaptability in these methods, recent work has shifted toward modeling the whole robot to construct {task-agnostic self-models \citep{kwiatkowski2019task,chen2022fully}.} They employed visual data with depth information to enable self-modeling of the overall morphology of a 4-degree-of-freedom robot. ~\citet{yang2024robotsdf} later adapted it to robots with higher degrees of freedom. \citet{chen2022fully}'s approach allows the robot to determine its spatial occupancy based on input joint configurations but cannot directly compute the kinematic information (e.g., position) of specific links. To address link-level modeling, ~\cite{yang2024robotsdf} developed multiple independent models for each robot link and integrated them into a comprehensive self-model. However, these methods require depth information to create self-models, which necessitates devices with higher cost compared to standard RGB cameras (e.g., depth cameras and LiDAR), and is hard to obtain in practice. In addition to visual data, inertial measurement unit (IMU) data can also be utilized for the self-modeling of robots~\citep{diaz2023machine}, which necessitates the attachment of numerous IMU sensors to robot links. Notably, despite the considerable advancements in robot self-modeling concerning morphology and kinematics, the aforementioned approaches have neglected the modeling of surface color, which is crucial for the visualization of robots in simulation environments and digital twin systems. When visualizing simulation results, a robot's 3D model with surface color is more conducive to understanding and more human-friendly. Currently, commonly used simulators are typically equipped with robot models with surface color, such as MuJoCo\citep{todorov2012mujoco}, Gazebo\citep{koenig2004design}, and Nvidia Isaac Sim\citep{liang2018gpu}. Thus, beyond morphology and kinematics, surface color modeling remains a significant challenge in achieving comprehensive robot self-modeling.

\begin{figure*}[tbh]
\centering
\includegraphics[width=.9\linewidth]{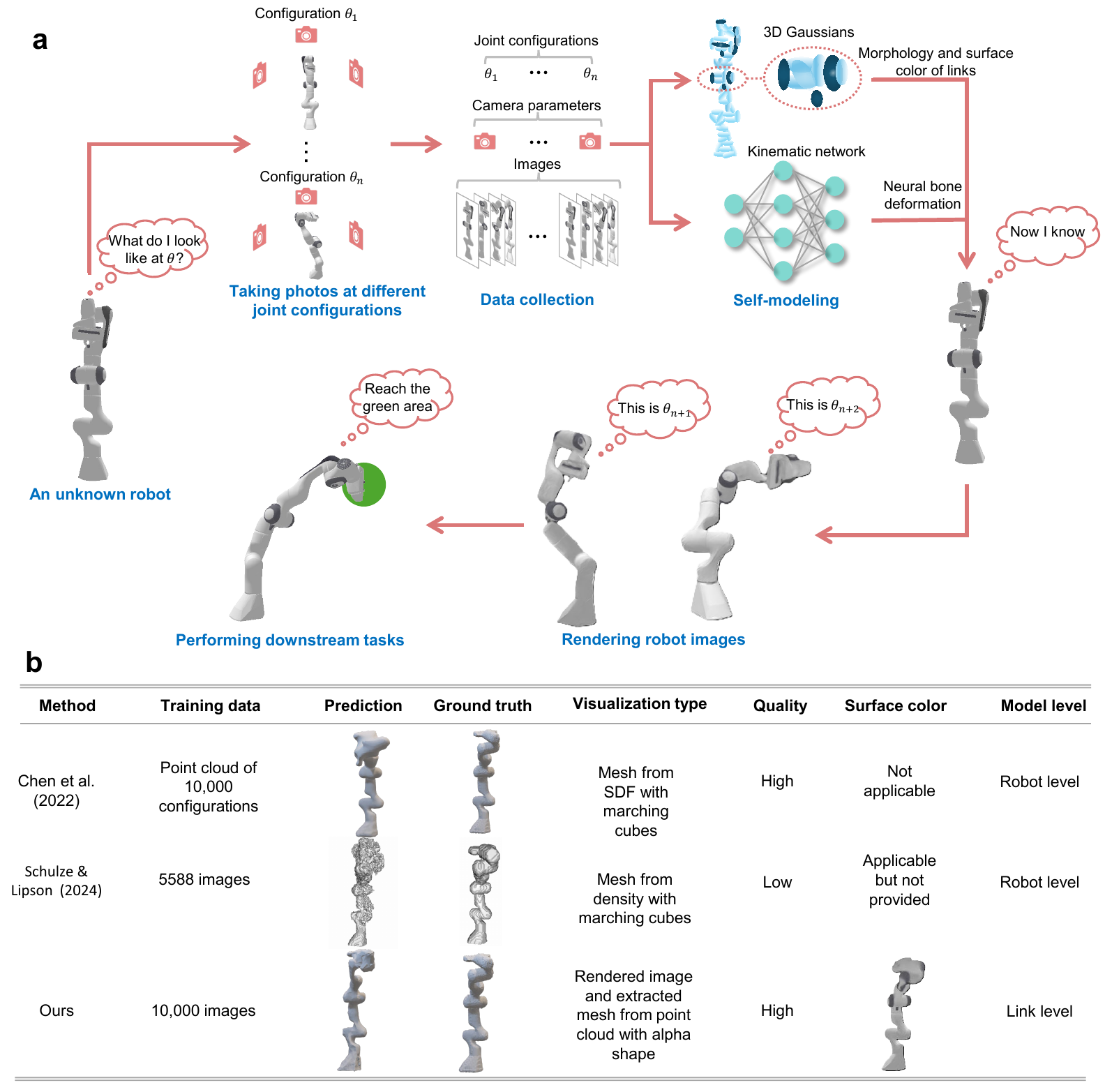}
\caption{Robot self-modeling from images. \textbf{a} Given a robot whose model is unknown, our method can build a self-model of the robot from multiple RGB images. We capture multi-view photos of the robot at different joint configurations, and train the 3D Gaussians and the kinematic network, which are responsible for representing the static morphology and surface color of robots and neural bone deformation, respectively. The well-trained model can be utilized to render robot images with the aid of 3D Gaussian splatting and perform downstream tasks. \textbf{b} Compared to the methods proposed by~\citet{chen2022fully} and \citet{schulze2024high}, {our method can learn a high-quality, surface color-aware, and link-level self-model}.}
\label{fig:overallfig}
\end{figure*}

Given the challenges and high costs in {equipment} for data acquisition in the previously discussed self-modeling methods and the importance of surface color modeling, a visual self-modeling approach utilizing only two-dimensional (2D) images presents a more efficient and cost-effective alternative. {However, the current 2D image-based self-modeling method suffers from poor morphology modeling quality and fails to model the link-based structure of robots \citep{schulze2024high}}. In this paper, we develop a scheme based on 3D Gaussian Splatting (3DGS)~\citep{kerbl3Dgaussians} and neural networks for modeling the robot's morphology, kinematics, and surface color. {3DGS is a technology for 3D scene reconstruction, which represents 3D scenes with a large number of optimizable, flexible, and expressive 3D Gaussian ellipsoids. Recent improvements equip it with the ability to real-time render scenes into images \citep{kerbl3Dgaussians}.} With the assistance of 3DGS, we can acquire 3D information, which is necessary for robot self-modeling from 2D images. In order to control 3D Gaussian ellipsoids to reflect the robot pose, we use a kinematic network to move them. An overview of our method is demonstrated in Figure~\ref{fig:overallfig}(a). To optimize the 3D Gaussians and kinematic network, the robot only needs to gather multi-view RGB images under various configurations, along with corresponding joint configuration data and camera parameters. The static morphology and surface color information of the robot's links are embedded in 3D Gaussians, each of which is an ellipsoid defined by its position, covariance matrix, color, and opacity. By clustering these 3D Gaussians, the bones of the robot model are constructed, with transformations of the bones determined by the kinematic network. By feeding the kinematic neural network with joint angles, we can utilize the well-trained self-model to describe the corresponding morphology, kinematics, and surface color of robots at the link level and render robot images from different perspectives with the aid of 3DGS. Besides, the self-model can be employed to directly control the robot to perform downstream tasks such as motion planning and inverse kinematics. As illustrated in Figure~\ref{fig:overallfig}(b), compared to previous methods, the proposed method {produces a high-quality and link-level self-model of robots}. 
Overall, the contributions of this paper are summarized as follows.
\begin{itemize}
    \item A novel robot self-modeling method based on 3D Gaussian Splatting is proposed, which enables efficient reconstruction from multi-view RGB images. {This is one of the first methods leveraging 3DGS in robot self-modeling.}
    \item The robot can be modeled at the link level. With the predefined link number of the robot, the self-model automatically identifies the spatial positions and kinematic connections of these links {based on the learnable and controllable neural bone architecture. This is the first method to use such architecture in robot self-modeling.}
    \item {Our proposed method achieves state-of-the-art performance, outperforming previous works of robot self-modeling in terms of geometric accuracy.}
    \item The self-model can be utilized to control the robot. Experiments show that the self-model can be utilized to control the robot to accomplish downstream tasks without additional training.
\end{itemize}

\section{Related work}
The main challenge of self-modeling robots from 2D visual input is how to reconstruct 3D models from 2D images or videos. Currently, there are two solutions: Neural Radiance Field \citep{NerF} and 3D Gaussian Splatting \citep{kerbl3Dgaussians}.

\subsection{3D reconstruction with NeRF}

Neural Radiance Field (NeRF) has shown remarkable results in reconstructing scenes from multiple-view photos. It uses {an} implicit function and volume rendering to represent a static scene. Spatial coordinates $x$ and view directions $d$ are mapped into corresponding density $\sigma$ and color $c$. All points along a ray that connects a pixel on the image and the camera's optical center are rendered into a {pixel on the image plane} by volume rendering, which is formulated as
\begin{equation}
        C(r) = \sum_{i=1}^{N}\exp\left(-\sum_{j < i}\sigma_j^r\delta_j^r\right)\left(1-\exp\left(-\sigma_i^r\delta_i^r\right)\right)c_i^r
\end{equation}
where $\delta_i^r$ is the distance between two sampled points along the ray, and $C(r)$ is the color on the rendered image. {However, during both training and rendering, it is necessary to sample hundreds or even thousands of points along each ray and perform a time-consuming network query for every single point to obtain its color and density, resulting in an extremely high computational cost.}
As the original NeRF is for static scenes only, \citet{Pumarola_2021_CVPR} proposed the first method to model dynamic scenes with NeRF. The key idea is to bend the ray and map the moved points to points on the original ray.

Some methods have been proposed to accelerate the training of NeRF~\citep{fang2022fast, mueller2022instant}. These works motivate researchers to use dynamic NeRF to reconstruct and animate the human body \citep{Noguchi_2021_ICCV} and animals \citep{yang2022banmo}. \citet{noguchi2022watch} combined NeRF with signed-distance functions (SDFs) to learn both the appearance and the structure of articulated objects, including robots and human bodies, by observing their movements from multiple views. They modeled articulated objects with several ellipsoids, and could repose articulated objects by transforming ellipsoids. \citet{uzolas2024template} extracted point clouds of articulated objects from pre-trained NeRF models and used Linear Blend Skinning (LBS) to learn dynamic NeRF and associated skeletal models jointly. Although these works achieved modeling robots from images or videos, they did not associate the robots' joints with the built morphology models. {In other words, they did not directly use joint angles as the primary deformation inputs to train their model and thus failed to model the robot kinematics described by joint angles. Controlling the model learned by their method requires more information, such as link length, link twist, and joint offset, in addition to joint angles.}

\subsection{3D reconstruction with 3DGS}

Though NeRF shows promising results in 3D reconstruction, it has a significant drawback. Training and rendering a high-quality scene reconstruction may take a long time. The 3DGS method is designed to solve this problem and can quickly complete training and learn high-quality models to present well-rendered images with a high peak signal-to-noise ratio. Similar to NeRF, researchers extended 3DGS to the reconstruction of dynamic 3D scenes~\citep{Wu_2024_CVPR,yang2023gs4d,liang2023gaufre, yang2024deformable, lin2024gaussian}. For example, \citet{hu2024gauhuman} and~\citet{qian20243dgs} proposed to model 3D human avatars with 3DGS. These works require a Skinned Multi-Person Linear (SMPL) model as a base human model~\citep{SMPL:2015}. Several other approaches aim to design controllable and editable models with 3DGS. \citet{yu2024cogs} used predefined masks to choose moving parts of the modeled scene. These masks are used to control the model, and only the chosen part can be controlled. \citet{huang2024sc} used control points to control the dynamic 3DGS model. Control points are automatically generated during training, which then control the movement of the whole 3DGS model with Linear Blend Skinning(LBS) \citep{sumner2007embedded}. \citet{zhang2024bags} proposed to use neural bones to control the 3DGS model, which was similar to the method proposed by \citet{yang2022banmo}. A neural bone is an ellipsoid parameterized by position and radius. Nevertheless, these works still did not use the joint values between links of an articulated object as inputs to the model, {and none of these works is designed for the purpose of robot self-modeling. }

\subsection{Robot self-modeling with 2D input}
{The methods mentioned above are primarily designed to model humans and animals, and are not intended for robot self-modeling. } Several studies have explored using 2D images for robot self-modeling \citep{hu2023teaching,marques2024visuo}. However, these approaches are either only applicable to simple robots with few degrees of freedom or limited to rendering images of the robot in various states, and do not succeed in constructing a comprehensive model of the robot's morphology. \citet{schulze2024high} was the first to solve the robot self-modeling problem using dynamic NeRF and associated the built model with input robot joints. However, NeRF-based self-modeling requires optimizing a dense multi-layer perceptron (MLP) for learning both morphology and kinematics, and uses volumetric ray-marching to render robot images. {The MLP expression makes the learned model difficult to be directly explained and edited. The ray-based volume rendering is also time-consuming.} As a result, NeRF-based self-modeling suffers from problems such as being difficult to balance training time and modeling quality. Another limitation of their approach is that they did not present results including estimated surface color, although the NeRF method they adopted is capable of rendering surface color. \citet{lou2024robogs} used a hybrid representation that combines 3DGS and mesh to create digital assets of robots. The model of robots can be controlled by the robot joints. However, real Denavit-Hartenberg parameters are required to control the model based on joints in training and inference stages, which requires human intervention and fails to model robot kinematics automatically. 

A concurrent work \citep{liu2024differentiablerobotrendering} also achieved modeling robots with RGB images only and maps joint values to the movement of the model. However, their method requires a known connection order of the robot's links to calculate the forward kinematics. Our method only requires the number of robot links.

In addition, \citet{labbe2021single} and \citet{lu2023image} also involve shape reconstruction and differentiable rendering. They focus on estimating the unknown joint angles of a robot given a known robot model and images of the robot. However, their work did not consider an unknown robot, which represents a fundamental difference from our approach. Our work focuses on learning an unknown model of a robot based on its images with joint annotations. A geometric model of the robot is also learned to visualize the learned robot's kinematics. While both papers and our work render images of robots, the purposes are different. Specifically, the rendered images in their work are generated based on estimated joint angles and the known robot model. The primary purpose of comparing these rendered images with real images is to quantify the accuracy of the estimated joint angles. In our work, image rendering is performed based on known joint angles and the learned geometric and kinematic models, for the purpose of evaluating the accuracy of the learned models. Even though \citet{lu2023image} learned a mesh reconstruction, they did not focus on the accuracy of the learned mesh reconstruction compared with a ground-truth robot mesh. Therefore, directly comparing the quality of rendered images of our work and theirs does not make sense because tasks and prior conditions are totally different.

\section{Method}\label{sec3}

\begin{figure*}[tbp]
    \centering
    \includegraphics[width=.9\linewidth]{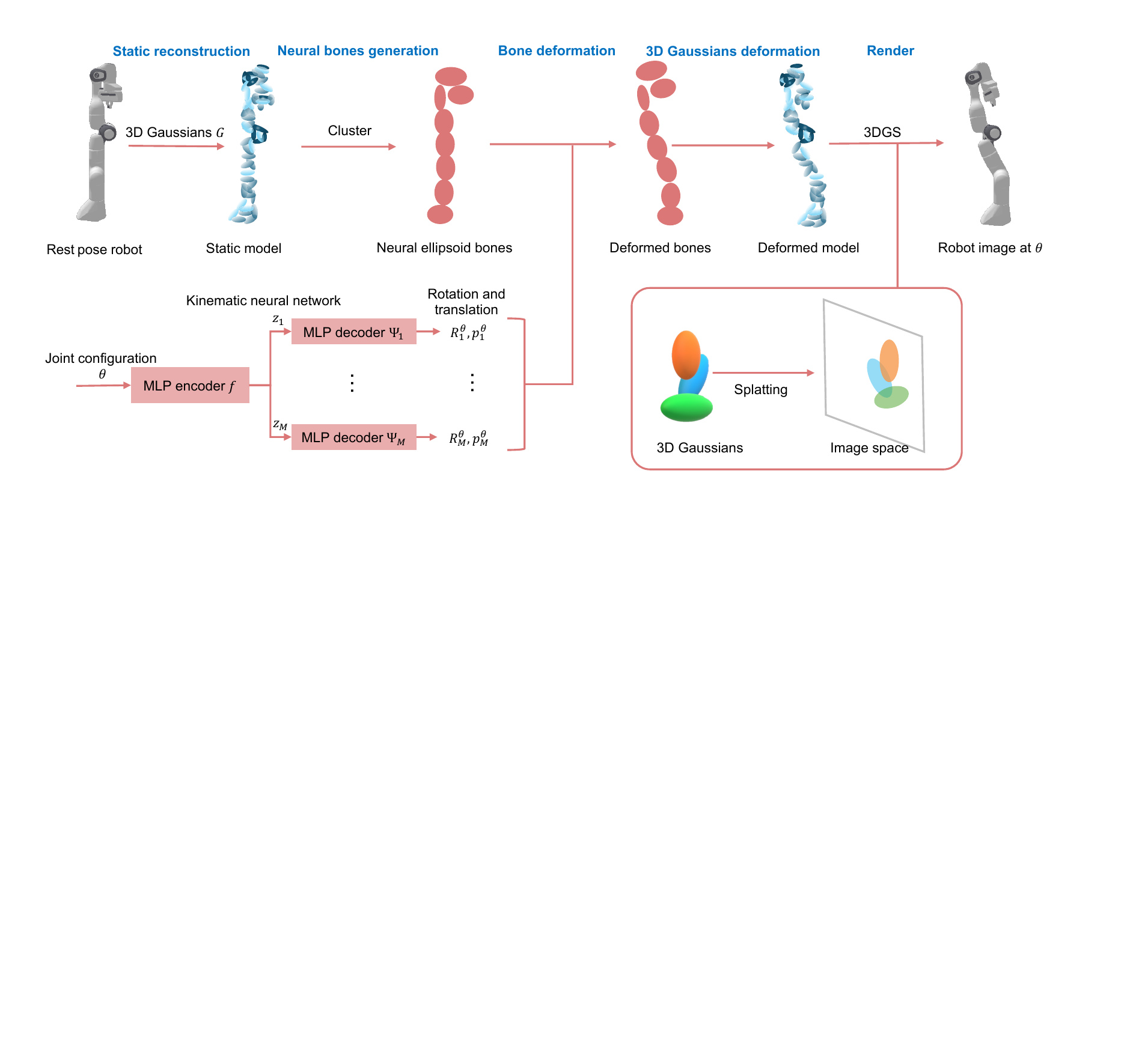}
    \caption{Overview of our method. We first train a static 3DGS self-model with images of the robot at zero pose from various views. Then, we initialize neural ellipsoid bones to model the robot kinematic chain. Using images of the robot at different joint configurations, we train a dynamic self-model. The neural bones are deformed, and the relative 3D Gaussian is controlled by LBS. Finally, the model is rendered into images and compared with the ground truth images.}
    \label{fig:method-demo}
\end{figure*}
We propose a novel robot self-modeling method that begins by modeling the robot in a rest pose, which we define as the canonical space. The robot is then transformed into various poses that form the transformed space. To implicitly represent the self-model of the robot, we utilize 3DGS. The model can be deformed and rendered into images, allowing us to visualize the robot's morphology from different perspectives. An overview of our method is provided in Figure~\ref{fig:method-demo}.

\subsection{Preliminaries}
3DGS uses 3D Gaussians to represent a static scene~\citep{kerbl3Dgaussians}. A 3D Gaussian is a small ellipsoid with feature parameters. Typically, a static 3DGS model consists of 10 thousand 3D Gaussians. Each 3D Gaussian is defined by a full 3D covariance matrix \(\Sigma\):
\begin{equation}\label{eqgx}
    G(x)\ =\ e^{-\frac{1}{2}{x}^T\Sigma^{-1}x}
\end{equation}
where $x\in \mathbb{R}^3$ denotes the position coordinate of a point in the ellipsoid. The covariance matrix \(\Sigma\) can be calculated using the rotation matrix \(R\) and the scaling matrix \(S\) as \(\Sigma=RSS^TR^T\). The rotation matrix is derived from a quaternion \(q\). Each Gaussian has an opacity value \(\sigma\) and a spherical harmonic coefficient $\xi$ to create a view-dependent color. When rendering images, 3D Gaussians are projected to a 2D plane. Given a viewing transformation \(W\), the covariance matrix \(\Sigma'\) in camera coordinates is given by
\begin{equation}\label{eqsigma}
    \Sigma'=(JW\Sigma W^TJ^T)
\end{equation}
where \(J\) is the Jacobian of the affine approximation of the projective transformation.
The color \(C\) of a pixel is computed by alpha blending 3D Gaussians that overlap on the pixel sorted by depth, which is given by
\begin{equation}\label{eqcolor}
    C=\sum_{i\in N}{c_i\alpha_i\prod_{j=1}^{i-1}{(1-\alpha_j)}}
\end{equation}
where \(N\) is the number of Gaussians overlapping the pixel, \(c_i\) is the color of Gaussian \(i\) calculated by spherical harmonic coefficient, \(\alpha \) is given by evaluating the projected Gaussian with covariance matrix \(\Sigma'\) and multiplied with its opacity value \(\sigma\). 3D Gaussian parameters {\(\mu, q, S, \sigma, \xi\)} are optimized during training, and the number of 3D Gaussians is adaptively controlled. {By representing scenes with a direct and explicit model of millions of 3D Gaussians, it can be rendered efficiently in a single pass via a splatting method—analogous to traditional graphics rasterization and highly optimized for modern GPUs—thereby achieving both training and rendering speeds.}

\subsection{Static reconstruction and neural bone}\label{static}
Our self-modeling method starts with modeling the robot in a rest pose. Given a set of images of a robot captured from different view angles, whose joints are all set to 0, we first reconstruct a static self-model of the robot. We then extract neural bones from the static self-model. 

Inspired by BANMo \citep{yang2022banmo} and BAGS \citep{zhang2024bags}, we use a set of neural bones to model the kinematic structure of the robot and control the transformation of the self-model. Each bone is represented by an ellipsoid, which is defined by a center point \(p\in\mathbb{R}^3\), a radii scaling matrix $r\in \mathbb{R}^3$, and a rotation expressed by quaternion $q_b \in \mathbb{R}^4$. We limit the maximum radii of each part with a hyperparameter. To initialize these neural bones, we use the K-means algorithm~\citep{10.5555/1283383.1283494} provided by Scikit-learn~\citep{scikit-learn} to cluster 3D Gaussians representing the static self-model. The number of clusters is the same as the number of neural bones. We use cluster centers to initialize center points $p_b$. The rotation of each bone is initialized to $(0, 0, 0, 1)$, and the radii are initialized to half of their maximum values. The number of bones can be either a custom value or one plus the number of joints of the robot. All parameters of neural ellipsoid bones are automatically optimized when training. We use a kinematic network to map joint angles into bone-specific transformation, which is explained below. In the experiment, we found that assuming that all neural bones have similar volumes can get the best result. In practice, we use eight bones for the robot with 7 degrees of freedom (DOF) in a simulation environment and seven for the 4-DOF robot in the real world. The 4-DOF robot we use has a large base, so we use multiple bones to represent it.

\subsection{Kinematic network}\label{kinematics}
The kinematic network predicts how the bones transform from the rest pose (canonical space) to a joint configuration pose (transformed space). The structure of our kinematic network is shown in Figure~\ref{fig:method-demo}. Assuming there are $M$ bones, the input joints are first encoded to bone-specific code $z_m$ with an encoder and then decoded to rotation and translation with the MLP decoder $\Psi$. We follow the idea of \citet{Tertikas2023CVPR} to build our encoder. The joint angle vector $\theta$ is first encoded with an MLP encoder $f$ and then decomposed into per-bone codes with a multi-head attention transformer $\tau$. We want each bone to focus on different parts of the encoded joint angle vector and learn features about its movement. The whole encoding process is as follows:

\begin{equation}\label{eqencoder}
    \{z_m\}_{m=1}^M =\tau(f(\theta)).
\end{equation}

Then the rotation $\Delta R_m^\theta$ and the {translation $\Delta p_m^\theta$ }of bone $m$ at joint configuration $\theta$ are decoded from the bone-specific code by the decoder $\Psi$. Using $R_m^c$ to represent the rotation matrix derived from the quaternion of bone $m$ at rest pose and using $p_m^c$ to represent the center point of bone $m$ at rest pose, the rotation matrix $R_m^\theta$ and position $p_m^\theta$ at joint configuration $\theta$ are given by
\begin{equation}\label{eqR}
    R_m^\theta=\Delta R_m^\theta R_m^c,
\end{equation}

\begin{equation}
    p_m^\theta=\Delta p_m^\theta + p_m^c.
\end{equation}

\subsection{3D Gaussian deformation}\label{deformation}
We use LBS~\citep{sumner2007embedded} to calculate the transformation of each 3D Gaussian based on the bone transformation that we obtain from the kinematic network. We use $w_{im}^\rightarrow$ to represent the skinning weight of 3D Gaussian $i$ with respect to bone $m$ when transforming from the rest pose to the joint configuration pose. Then, the center point $\mu_i^\theta$ and the rotation matrix $R_i^\theta$ derived from the quaternion of the deformed 3D Gaussian $i$ at joint configuration $\theta$ can be computed by

\begin{equation} \label{mucalforward}
    \mu_i^\theta=\sum_{m\in M}{w_{im}^\rightarrow(\Delta R_m^\theta(\mu_i^c-p_m^c)+p_m^c+\Delta p_m^\theta)},
\end{equation}

\begin{equation}\label{rcalforward}
    R_i^\theta=\sum_{m\in M}{w_{im}^\rightarrow{\Delta R}_m^\theta}R_i^c
\end{equation}
where $\mu_i^c$ is the center point of 3D Gaussian $i$ at the rest pose, and $R_i^c$ is the rotation matrix of 3D Gaussian $i$ at the rest pose.
To simplify the expression of the transformation, we use $\mu_i^\theta=\Theta^{\theta,\rightarrow}(\mu_i^c)$ to represent \eqref{mucalforward}. Similarly, we also define an inverse transformation  $\mu_i^c=\Theta^{\theta,\leftarrow}(\mu_i^\theta)$ that is calculated as follows:

\begin{equation} \label{mucalbackward}
    \mu_i^c=\sum_{m\in M}{w_{im}^\leftarrow\left(\left(\Delta R_m^\theta\right)^{-1}\left(\mu_i^\theta-p_m^\theta\right)+p_m^\theta-\Delta p_m^\theta\right)}.
\end{equation}

The calculation of skinning weights $w_{im}$ includes two parts. The original skinning weight $w_{im}^o$ is determined by the Mahalanobis distance between 3D Gaussian $i$ and bone $m$, as mentioned by \citet{yang2021lasr}:
\begin{equation}
    w_{im}^o=(\mu_i-p_m)^TR_m^Tr_iR_m(\mu_i-p_m)
\end{equation}
where $r_i$ is a $3 \times 3$ diagonal radius matrix of the 3D Gaussian $i$. When computing the weights for the transformation from the rest pose to the joint configuration pose, the distance is computed with the center points and rotation matrices of 3D Gaussian $i$ and bone $m$ at the rest pose. The weights for the inverse transformation are computed with the same values at the joint configuration pose.

Similar to the approach proposed by~\citet{yang2022banmo}, delta skinning weights are used for fine geometry. Delta skinning weight $w_{im}^\Delta$ is given by a weight MLP $\omega$, which takes the joint angle vector and the center position of a 3D Gaussian as input, and outputs a delta weight for each bone:
\begin{equation} \label{dsk}
    w_{im}^\Delta=\omega(\mu_i, \theta)
\end{equation}
where $\theta$ is the joint angle vector. To calculate the forward transformation weights, $\theta$ is the rest pose joint angle vector and can be expressed as $\theta^c$. To calculate the inverse transformation weights, $\theta$ denotes the joint configuration vector and can be expressed as $\theta^t$.

\begin{figure}
    \centering
    \includegraphics[width=1\linewidth]{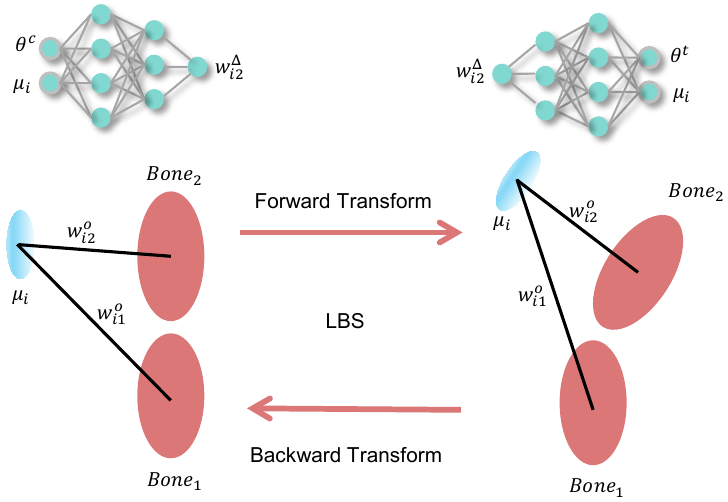}
    \caption{Illustration of 3D Gaussian deformation. We use one 3D Gaussian and two neural bones as an example. The deformation of the 3D Gaussian is a weighted sum of the transformation of the two bones. The weight is divided into two parts. One is based on the distance between 3D Gaussian and neural bones, and the other is predicted by a weight MLP.}
    \label{fig:transform}
\end{figure}

The final skinning weight is the sum of the two parts of weights and normalized by a softmax function:
\begin{equation} \label{skining weight}
    w_{im}=\text{softmax}(-s(w_{im}^o + w_{im}^\Delta))
\end{equation}
where $s$ is a learnable scaling factor. A demonstration of the 3D Gaussian deformation is illustrated in Figure~\ref{fig:transform}.

With the 3D Gaussian deformation method, we can control the static self-model. Combining the static self-model and the deformation method, we build a dynamic and controllable self-model of a robot and obtain its morphology, kinematics, and surface color.

\subsection{Optimization}\label{optimization}

We optimize 3D Gaussians, parameters of the kinematic network, parameters of all neural bones, parameters of the weight MLP, and the scaling factor for the skinning weight. We observed that training 3D Gaussians, the kinematic network, and neural bones simultaneously often led to convergence issues. To address these challenges, after training the static 3DGS self-model, we freeze the parameters of all 3D Gaussians and focus solely on training the kinematic network. Additionally, due to the robot's high degrees of freedom, it exhibits a wide range of postural variations with random joint configurations, and the kinematic network struggles to train effectively. Therefore, during the initial training phase, we restrict the robot's joint motions to a limited range by using the small-range dataset and gradually relax these constraints over time. This approach enables the model to learn kinematics progressively, starting from simpler to more complex instances. Once the range of motion reaches its maximum, we resume joint training of the 3D Gaussian parameters and the kinematic network. In total, we trained the self-model in over 600,000 steps with a joint limitation of $\frac{\pi}{2}$. Another 300,000 steps are required to train the self-model without joint limitation.

\subsection{Loss functions}
Our optimization objective $L$ is the sum of several loss functions. Our loss functions can be separated into four parts and used to train different parts of our method, including the image part, the 3D Gaussian part, the skinning weight part, and the neural bone part. For the image part, we use the same image reconstruction loss $L_{render}$ in 3DGS, which is a combination of the $L_1$ loss and a D-SSIM term to compare the input with the rendered image:
\begin{equation}\label{renderloss}
    L_{render}=(1-\lambda)L_1+\lambda L_{D-SSIM}.
\end{equation}
We also use a mask loss $L_{mask}$ that compares the foreground mask of the input image and the rendered foreground mask, which is calculated as the following $L_1$ loss:
\begin{equation}
    L_{mask} = ||I_m- \hat{I}_m||_1
\end{equation}
where $I_m$ is the ground truth mask and $\hat{I}_m$ is the rendered mask.

Then, for the 3D Gaussian part, following the previous work \citep{prokudin2023dynamic}, we use the as-isometric-as-possible constrain \citep{kilian2007geometric} to restrict the distance between neighboring 3D Gaussians to keep similar before and after deformation:
\begin{equation}
    L_{isopos}=\sum_{i=1}^{N}{\sum_{k\in\mathcal{N}}\frac{\left|d(\mu_i^c-\mu_k^c)-d(\mu_i^\theta-\mu_k^\theta)\right|}{KN}}
\end{equation}
where $N$ is the total number of 3D Gaussians and $\mathcal{N}$ is the set of $K$ nearest neighborhoods of 3D Gaussian $i$.

We also use the local-rigid loss $L_{rigid}$ and the rotational loss $L_{rot}$~\citep{luiten2024dynamic} to constrain the kinematic network. Each of these losses is assigned a weight determined by the distance between 3D Gaussians:
\begin{equation}
    w_{i,k}=\exp(-\lambda_w\|\mu_k^c-\mu_i^c\|_2^2).
\end{equation}
Then, the local-rigid loss $L_{rigid}$ is defined as follows:
\begin{equation}
   L_{rigid}=\sum_{i=1}^{N}{\sum_{k\in\mathcal{N}}\frac{w_{i,k}\|(\mu_k^\theta-\mu_i^\theta)-R_i^\theta{(R_i^c)}^{-1}(\mu_k^c-\mu_i^c)\|_2}{KN} }.
\end{equation}
This loss forces all nearby Gaussians of Gaussian $i$ to move in the same way as it. Similarly, the rotational loss is given by
\begin{equation}
    L_{rot}=\sum_{i=1}^{N}{\sum_{k\in\mathcal{N}}\frac{w_{i,k}\|R_k^\theta{(R_k^c)}^{-1}-R_i^\theta{(R_i^c)}^{-1}\|_2}{KN} }.
\end{equation}
This loss ensures that all nearby Gaussians of Gaussian $i$ have the same rotation at different joint configurations. One could refer to \citet{luiten2024dynamic} for more details about these two loss functions.

For the skinning weight part, we use a 3D cycle loss~\citep{yang2022banmo} to encourage the delta skinning weight that is calculated by \eqref{dsk} under different joint configurations:
\begin{equation}
    L_{cycle}=\frac{1}{N}\sum_{i=1}^N{\|\Theta^{\theta,\leftarrow}(\Theta^{\theta,\rightarrow}(\mu_i^c))-\mu_i^c\|_2^2}.
\end{equation}

A smooth loss is also employed to encourage neighboring 3D Gaussians to have similar skinning weights:
\begin{equation}
        L_{smooth}=\frac{1}{KN}\sum_{i=1}^N{\sum_{k\in\mathcal{N}}}|w_i-w_k|.    
\end{equation}

Finally, we use a bone mask loss $L_{bone}$, a bone center loss $L_{center}$, and a bone volume loss $L_{volume}$ to train neural bones. We treat each neural bone as a 3D Gaussian and render it into a binary image. The bone mask loss is given by the $L_1$ loss of the bone binary image and the robot's mask. To discourage neural bones from being concentrated in a small region of the robot, inspired by~\cite{noguchi2022watch}, we penalize small distances between their centers:
\begin{equation}
    L_{center}=\frac{1}{M(M-1)}\sum_{m\neq n}{\exp(-\frac{\left||p_m-p_n|\right|_2^2}{2\sigma^2})}
\end{equation}
where $\sigma$ controls the scale of the distances to be regularized. We also encourage that bones have comparable volume by the bone volume loss:

\begin{equation}
    L_{volume}=\frac{2}{M(M-1)}\sum_{j=1}^{M}{\sum_{k=1}^{j-1}{\left|\mathcal{V}(r_j)-\mathcal{V}(r_k)\right|}}
\end{equation}

where $\mathcal{V}(\cdot)$ is the volume of the neural ellipsoid bone calculated by the radius scaling matrices.

The final loss is a weighted sum of all loss functions mentioned above:
\begin{equation}
    \begin{aligned}
        L=&\lambda_1 L_{render}+\lambda_2 L_{mask} 
        \\&+\lambda_3 L_{isopos}+\lambda_4 L_{rigid} +\lambda_5 L_{rot}
        \\&+ \lambda_6 L_{cycle} + \lambda_7L_{smooth}
        \\& + \lambda_8 L_{bone} + \lambda_9 L_{center} +\lambda_{10} L_{volume}
    \end{aligned}
\end{equation}
where $\lambda_i$ (with $i = 1,2,\cdots,10$) is the weight coefficient. The first, second, third, and fourth lines represent the losses for images, 3D Gaussians, skinning weights, and neural bones, respectively.

\subsection{Implementation details}
For the static reconstruction, we randomly generated a point cloud with 50,000 points as the base of 3D Gaussians. We set the densification interval of 3DGS to 200 to constrain the total number of 3D Gaussians during optimization. 

We used an MLP with three layers of 128 dimensions for the kinematic network to encode the input joint values. We used a transformer encoder with four heads and two layers to encode the part-specific feature. The MLP decoder has eight layers with 128 dimensions. We used quaternions to represent rotation. For the weight MLP in 3D Gaussian deformation, we used an MLP with six layers of 256 dimensions.

In terms of optimization, we used the same setup with the original implementation of 3DGS \citep{kerbl3Dgaussians} to train 3D Gaussians. We used AdamW \citep{loshchilov2019decoupledweightdecayregularization} to optimize the rest part of our model. We set the learning rate to $1\times 10^{-4}$ and use a cosine annealing schedule~\citep{loshchilov2017sgdrstochasticgradientdescent}. The weight coefficients for all loss functions were empirically set as follows: $\lambda_1 = 1, \lambda_2 = 0.1, \lambda_3 = 1, \lambda_4=0.1, \lambda_5=0.1, \lambda_6=1, \lambda_7=10, \lambda_8=0.1, \lambda_9=0.1, \lambda_{10}=1$. The training process of 600,000 steps took about 16 hours on a single Nvidia RTX 3090 GPU to learn a self-model with joint limitation, and the additional 300,000 steps required to train a self-model without joint limitation took about 8 hours. {Besides, we replaced the 3DGS backbone in our method with NeRF, and the NeRF version method required approximately two days of training on a single RTX 3090 GPU, yet produced significantly inferior rendering quality compared to our method.}

\section{Results}\label{sec2}

We validated the proposed self-modeling method in a simulation environment using a 7-DOF Franka robot and in an experimental setting with a 4-DOF OpenManipulator robot. Subsequently, we demonstrated the potential of our method in downstream tasks through several examples of motion planning and inverse kinematics. All of our experiments are conducted on an Nvidia RTX 3090 GPU, and the per-image rendering time is approximately 0.08 seconds.

\begin{figure*}[tbhp]
    \centering
    \includegraphics[width=0.9\textwidth]{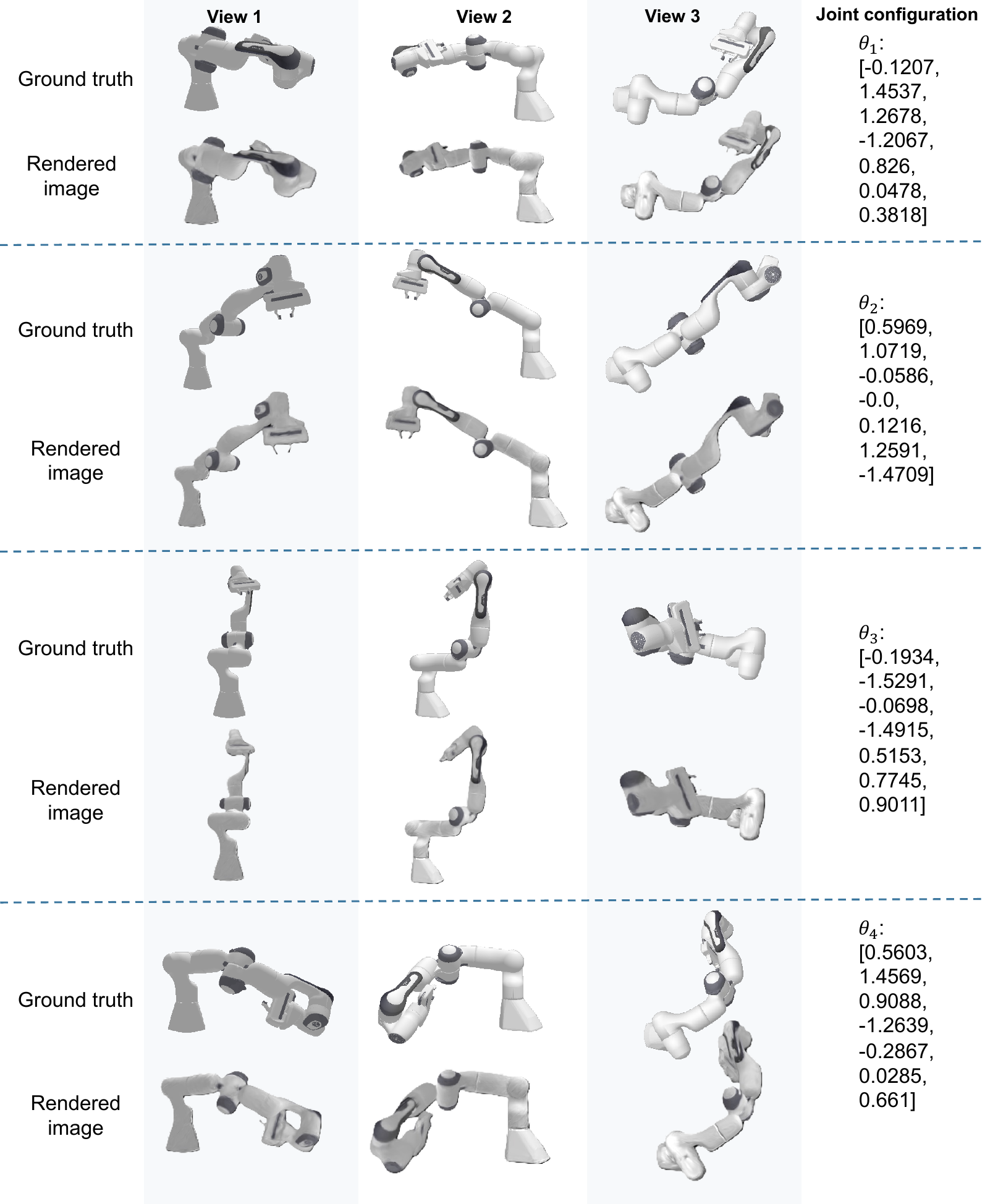}
    \caption{Self-modeling result and ground truth in the simulation environment. Given a random joint configuration, we rendered the self-model to images from 3 different view angles. The joint configurations are shown in radians.}
    \label{fig:simulation}
\end{figure*}

\subsection{Simulation}

To train the model, we followed the steps below to generate $400\times400$ resolution images of the 7-DOF Franka robot in the Pybullet simulator~\citep{coumans2021}. First, all joints of the robot were set to 0 radians, i.e., $\theta = [0, 0, 0, 0, 0, 0, 0]^T$ rad. A total of 100 images of the stationary robot were captured from various random view angles. Next, we actuated the robot by selecting different joint angles. Initially, we constrained all joint angles within $[-\frac{\pi}{6},\frac{\pi}{6}]$ and then expanded them to $[-\frac{\pi}{3},\frac{\pi}{3}]$. Finally, we limited joint movement to the range of $[-\frac{\pi}{2},\frac{\pi}{2}]$. For each range, we randomly sampled seven joint angles within it and moved the simulated robot to the sampled pose. When the robot was stable, we recorded the actual joint angles attained, in case some randomly generated target configurations might be kinematically unreachable, {as some configurations would cause self-collision.} In most of our simulations, we maintained the joint angles in the final range, as this range allows the robot to perform a large number of tasks and leads to better image rendering quality while simplifying the self-model training process. We took one image for each robot pose from a random view angle, and the camera parameters for each image were calculated and saved during the image generation process. A total of 10,000 images were generated: 500 images with joints constrained in $[-\frac{\pi}{6},\frac{\pi}{6}]$, another 500 images with joints constrained in $[-\frac{\pi}{3},\frac{\pi}{3}]$, and the remaining 9,000 images with joints constrained in $[-\frac{\pi}{2},\frac{\pi}{2}]$. Additionally, we captured 9,000 images without joint constraints to demonstrate that our method can handle this situation effectively.

\begin{figure*}[tbh]
    \centering
    \includegraphics[width=.8\textwidth]{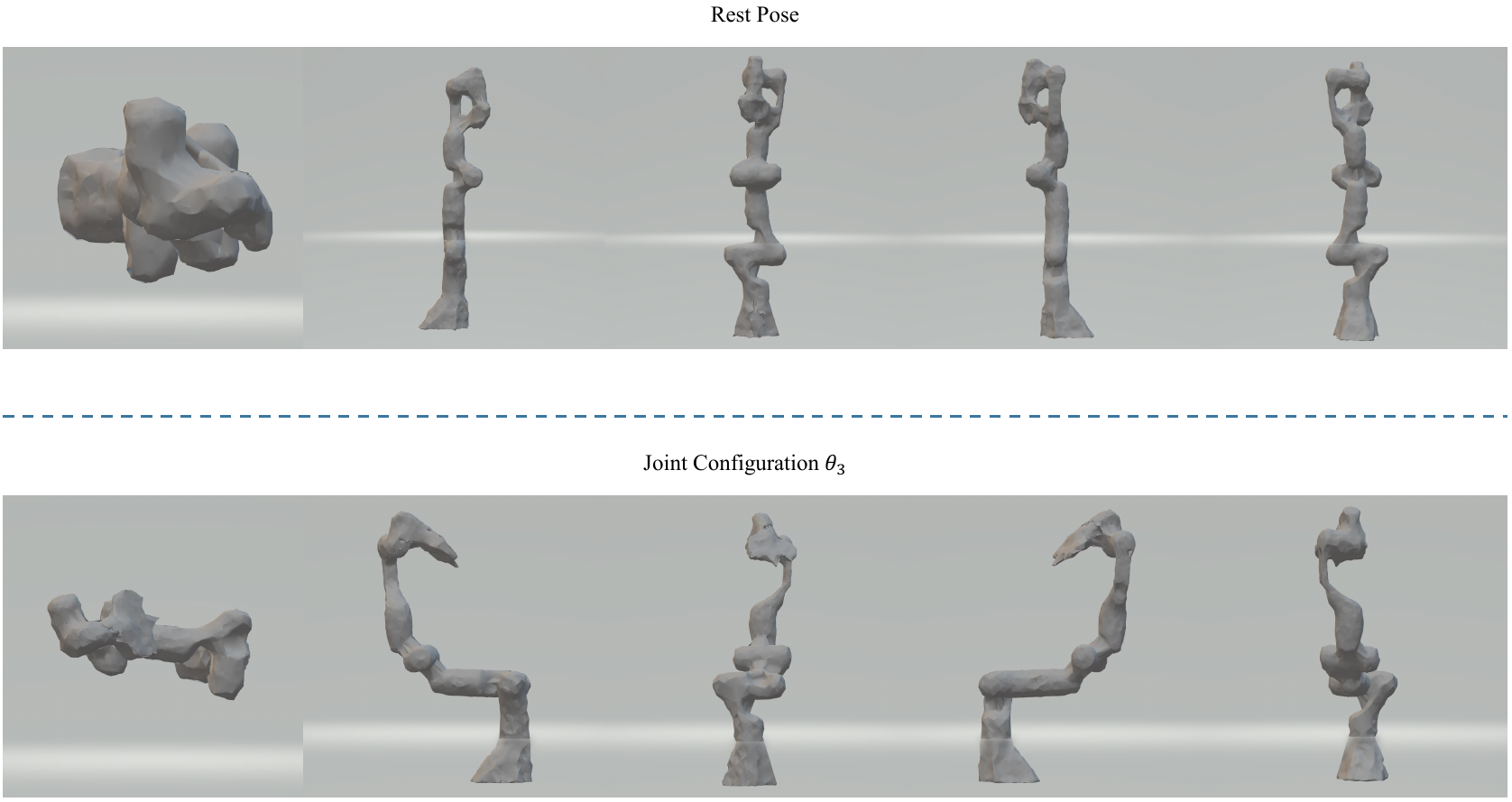}
    \caption{Mesh extracted from the point could formed by 3D Gaussians. We illustrate five views of the extracted meshes. They are one top-down view along with four orthogonal views: front, back, left, and right. }
    \label{fig:mesh_limit}
\end{figure*}

The results in the simulation environment are shown in Figure~\ref{fig:simulation}. Four different joint configurations were randomly generated to control the self-model, and the self-model is rendered into images from three different views, including two horizontal views and a top view. Comparing the rendered images with ground truth images, it is obvious that the self-model built with our method correctly reflects the robot's morphology, kinematics and surface color under different joint configurations. Besides the rendered images, we treated centers of 3D Gaussians as point clouds, and extracted a mesh with alpha shapes \citep{1056714} {because it is the most direct method for converting a point cloud into a mesh}. {We set the parameter $\alpha$ to 0.022}. We show meshes of the robot at the rest pose and joint configuration $\theta_3$ in Figure \ref{fig:mesh_limit} from five views. As we employed the original version of the 3DGS method, which inherently lacks the capability to directly generate a mesh, the quality of the meshes extracted using our approach is subject to degradation.

\begin{figure*}[tbh]
    \centering
    \includegraphics[width=.8\textwidth]{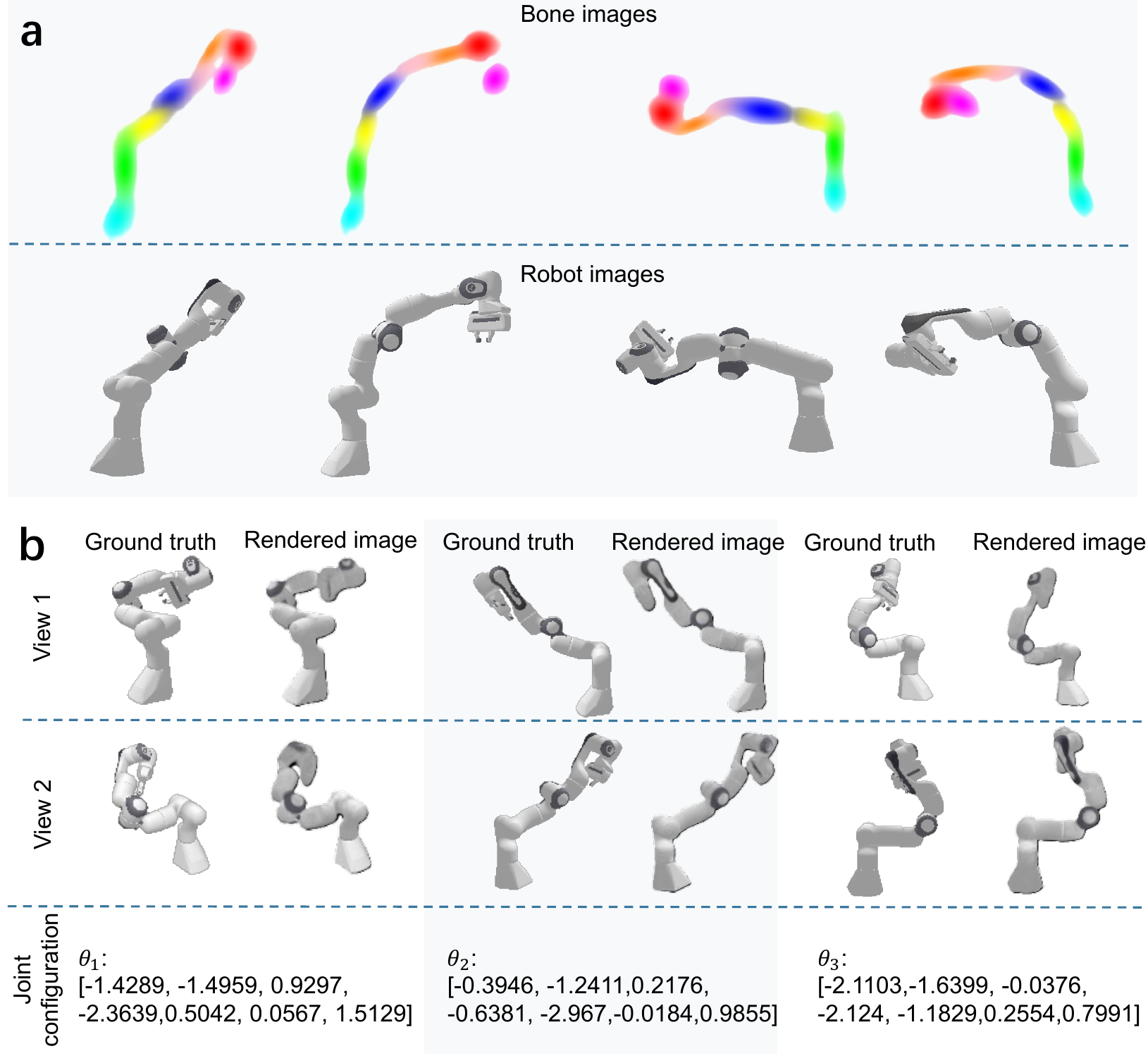}
    \caption{\textbf{a} Bone images in different joint configurations and rendered images at the same configuration. Every neural bone is assigned a color randomly and rendered to images in the same way as in 3DGS. These neural bones form the kinematic link of the robot. The bone images correspond to the robot images at each joint configuration. \textbf{b} Simulation results when joint ranges are not limited. The reconstruction quality is worse than the case with limited joint ranges. However, the self-model still gives reliable results on how the robot moves at different joint configurations.}
    \label{fig:bone}
\end{figure*}

As illustrated in Figure~\ref{fig:bone}(a), we assigned a distinct color to each neural bone and rendered them into an image, demonstrating the self-model's ability to learn the robot's kinematics. Eight neural bones were used in the self-model. They are not constrained to be linked together, as each bone is controlled independently and positioned according to its intended location under each joint configuration of the robot. The order of connection and the size of each bone are automatically determined by the self-model.

\begin{table*}[h]
\small\sf\centering
\caption{Quantitative results in the simulation environment with joint limitations.}
    \centering
    \begin{tabular}{c c c c c c} \hline 
        \toprule
         Metric&  Configuration $\theta_1$&  Configuration $\theta_2$&  Configuration $\theta_3$&  Configuration $\theta_4$& Test Set\\
         \midrule
         PSNR$\uparrow$&  $29.56$&  $29.30$&  $28.70$&  $30.30$& $31.22$\\ 
         SSIM$\uparrow$&  $0.984$&  $0.981$&  $0.981$&  $0.985$& $0.988$\\ 
         LPIPS$\downarrow$&  $0.0159$&  $0.0175$&  $0.0171$&  $0.0152$& $0.0137$\\
         Chamfer distance$\downarrow$& $2.527 \times10^{-4}$ & $2.380 \times 10^{-4}$ & $2.483 \times 10^{-4}$ & $2.414 \times 10^{-4}$ & $2.761 \times 10^{-4}$\\
         \bottomrule
    \end{tabular}
    \label{tab:simulate_label}
\end{table*}

The quantitative results of the above four different joint configurations are presented in Table~\ref{tab:simulate_label}. The metrics we used are peak signal-to-noise ratio (PSNR)~\citep{hore2010image}, structural similarity (SSIM)~\citep{odena2017conditional,wang2004image}, and learned perceptual image patch similarity (LPIPS)~\citep{zhang2018unreasonable}, which are commonly used to evaluate the quality of synthetic images. The quantitative results for each joint configuration were calculated using seven images, consisting of six horizontal views around the robot and one top view. We also validated the proposed method in a test set containing 2,000 images captured from random view angles around the robot, with varying joint configurations. The results of the test set indicate that our self-model can correctly synthesize images of the robot at different joint configurations and view angles. It is worth noting that larger changes in joint configurations make it more challenging to obtain an accurate self-model, and the test set includes joint configurations with minor changes. Consequently, the results for the 4 example joint configurations are worse than those of the test set. We also computed the Chamfer distance between the extracted mesh and the ground truth mesh for the four demonstrated joint configurations. We uniformly sampled 10,000 points on the mesh, {and used the PyTorch3D \citep{ravi2020pytorch3d} library to calculate the Chamfer distance}. The calculation of Chamfer distance can be formulated as:

\begin{equation}
    \begin{aligned}
            CD(S_1, S_2) &= \frac{1}{|S_1|} \sum_{x \in S_1} \min_{y \in S_2} \left\| x - y \right\| \\& + \frac{1}{|S_2|} \sum_{y \in S_2} \min_{x \in S_1} \left\| y - x \right\|
    \end{aligned}
\end{equation}
where $S_1$ and $S_2$ are two point clouds sampled from two meshes.

As noted previously, our method does not provide a direct 3D representation. Consequently, meshes extracted from the model may not accurately reflect its modeling quality, which reduces the utility of using the Chamfer distance metric for evaluation. 

\begin{figure*}[tbh]
    \centering
    \includegraphics[width=.9\textwidth]{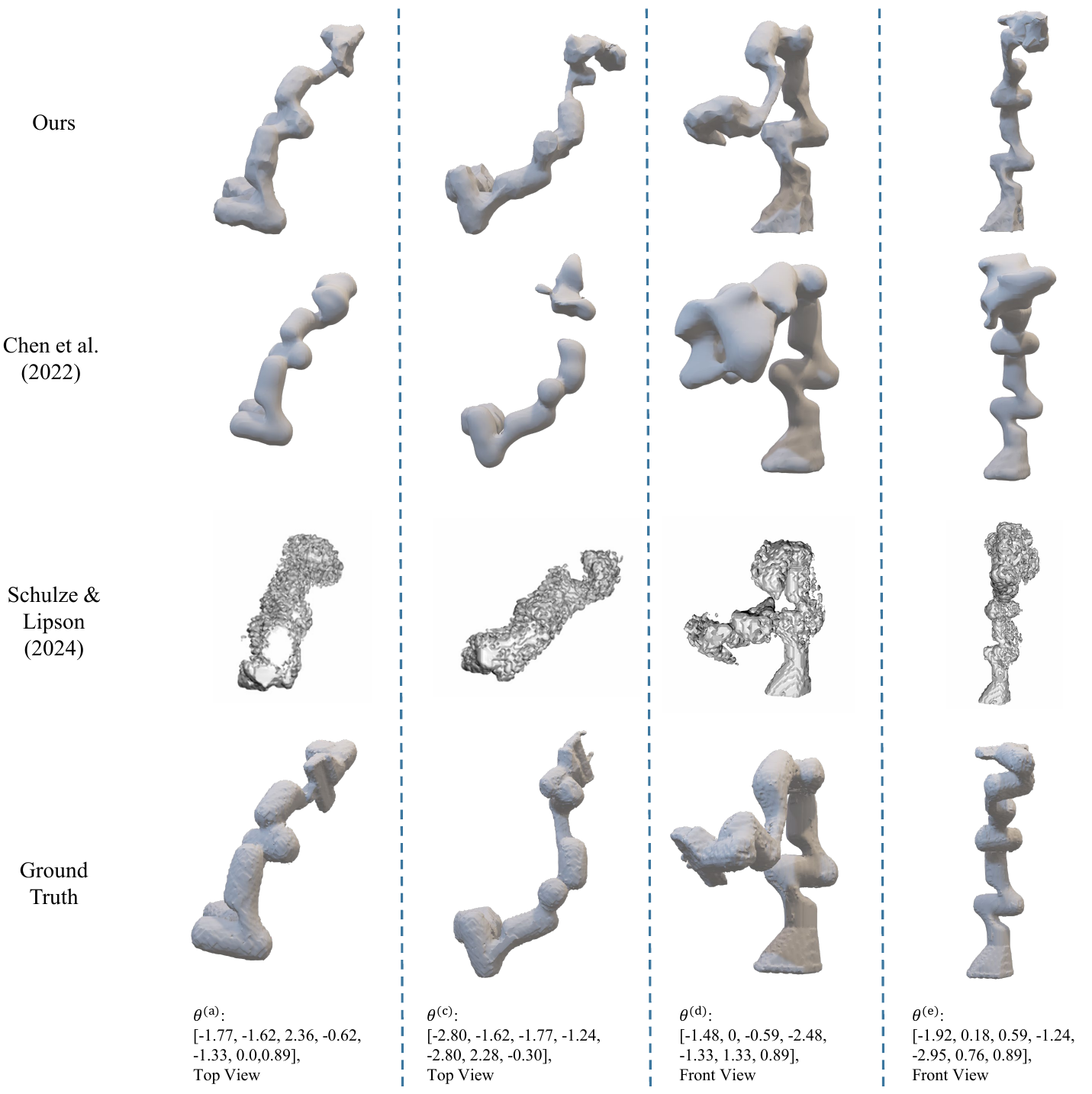}
    \caption{Mesh comparison of our method, the method in \citet{chen2022fully}, and the method in \citet{schulze2024high}. Meshes provided by our method are more easily identifiable by human perception.}
    \label{fig:mesh_compare}
\end{figure*}

\begin{figure}[tbh]
    \centering
    \includegraphics[width=.4\textwidth]{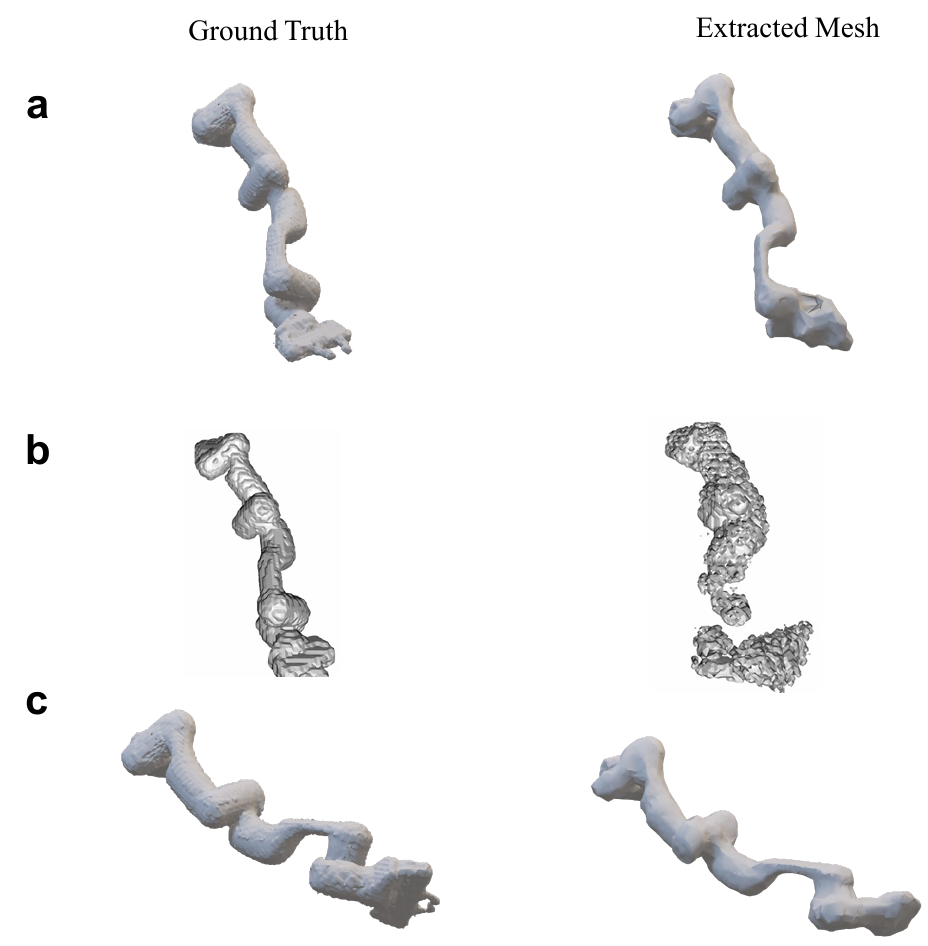}
    \caption{\textbf{a} Ground truth we rendered and meshes extracted under the changed joint configuration. \textbf{b} Images of ground truth and extracted mesh provided by \citet{schulze2024high}. \textbf{c} Ground truth we rendered and mesh extracted under the original provided joint configuration.}
    \label{fig:joint_b}
\end{figure}

We also present results without joint limits when posing the robot, as shown in Figure~\ref{fig:bone}(b). For each joint configuration, we display two images from different random view angles, with at least one of the joint angles exceeding $\frac{\pi}{2}$. Self-modeling with a larger range of joint angles is significantly more challenging, leading to results that are worse than those obtained within a smaller range. The robot's end effector is not precisely reconstructed due to its position at the end of the kinematic chain and the complexity of its movement. However, other parts of the robot are still accurately modeled. Furthermore, we conducted a comparison of the meshes generated by our method, the method proposed by \citet{chen2022fully}, and the method proposed by \citet{schulze2024high}. These meshes are illustrated in Figure~\ref{fig:mesh_compare}. We also calculated the Chamfer distance as listed in Table~\ref{tab:chamfer}. {Note that the chamfer distance of the method proposed by \citet{schulze2024high} is directly taken from their paper.} We extracted meshes of the robot at the same joint configurations presented by \citet{schulze2024high}, {and we used the same annotations}. However, $\theta^{(b)}$ was omitted for the sake of rigor, because a significant discrepancy between our model's results and the provided ground truth is observed at only this joint configuration {and we assumed the value they provided is incorrect. After changing the second angle value from a negative to a positive value (from $[-0.89,1.26,2.36,-0.62,-1.92,2.47,-1.77]$ to $[-0.89,-1.26,2.36,-0.62,-1.92,2.47,-1.77]$)}, the results exhibited closer agreement with the ground truth. We illustrated this in Figure \ref{fig:joint_b}. We implemented the method in \citet{chen2022fully} based on the code released by them to generate meshes and adapted it to the 7-DOF robot used in our simulation by simply changing the input dimension to 7. The training took about 2 days on a single Nvidia RTX 3090 GPU. As for the method in \cite{schulze2024high}, since the robot and joint angles we used are the same, we directly compared the results in their paper. Compared to the method proposed by~\citet{schulze2024high}, which uses a NeRF-based method for self-modeling, and the method proposed by \citet{chen2022fully}, which uses an SDF-based method and requires depth information, our method reconstructs the robot's shape more precisely, allowing easier identification of the robot's surface color and morphology. {Nevertheless, one may notice that in configuration $\theta^{(c)}$, the method of \citet{chen2022fully} attains the lowest chamfer distance. A possible reason is that as we sample 10,000 points on the mesh, the mesh quality for the links near the robot's base is very high, and they account for a large portion of the model. However, near the end of the robot, the disconnected section has a ground truth mesh that is inherently thin, resulting in fewer collected data points for that area. Consequently, when the metric is averaged, this leads to a lower overall Chamfer distance.}

\subsection{Physical experiment}

We utilized a smartphone and an Omnivision OV5640 color sensor attached to the end effector of a robot to capture images of the OpenManipulator robot. All images were resized to a resolution of $640 \times 480$ while maintaining the aspect ratio. Following a procedure similar to that in the simulation, we first used both cameras to capture images with the robot in its zero pose. The smartphone was free to capture images from various view angles, whereas the color sensor could only capture images from fixed angles. To simplify data collection, we captured images in a forward-facing manner instead of a $360^\circ$ inward approach, {which means we used multiple fixed cameras positioned within a certain angular range in front of the robot to capture the images of the robot}. Notably, multiple fixed cameras could be used to accelerate the data collection process. Camera parameters were computed using structure-from-motion (SfM)~\citep{schoenberger2016sfm} based on the static scene. Subsequently, we randomly posed the robot and used only the color sensor to capture images from the calculated view angles. We restricted the joint angles in $[-\frac{\pi}{6},\frac{\pi}{6}]$ to capture 100 images, then posed the robot without joint limitations to capture additional 1,000 images. We employed the segment anything model (SAM)~\citep{kirillov2023segany} to generate masks and segment the robot from the background.

\begin{table*}[h]
\small\sf\centering
\caption{Comparison of Chamfer distances without joint limitations.}
    \centering
    \begin{tabular}{c c c c c} \hline 
        \toprule
         Method & Configuration $\theta^{(a)}$&  Configuration $\theta^{(c)}$&  Configuration $\theta^{(d)}$&  Configuration $\theta^{(e)}$\\
         \midrule
         Ours&$3.6\times10^{-4}$&  $4.4\times10^{-3}$&  $4.5\times10^{-4}$&  $3.4\times10^{-4}$\\ 
         \citet{schulze2024high} &$1.7 \times 10^{-2}$&  $5.3 \times 10^{-2}$&  $1.3\times10^{-2}$&  $1.3\times10^{-2}$\\ 
          \citet{chen2022fully} & $4.3\times10^{-4}$ &  $5.8\times10^{-4}$&  $2.3\times10^{-3}$&  $1.7\times10^{-3}$\\ 
         \bottomrule
    \end{tabular}
    \label{tab:chamfer}
\end{table*}

\begin{figure*}[tbhp]
    \centering
    \includegraphics[width=.9\linewidth]{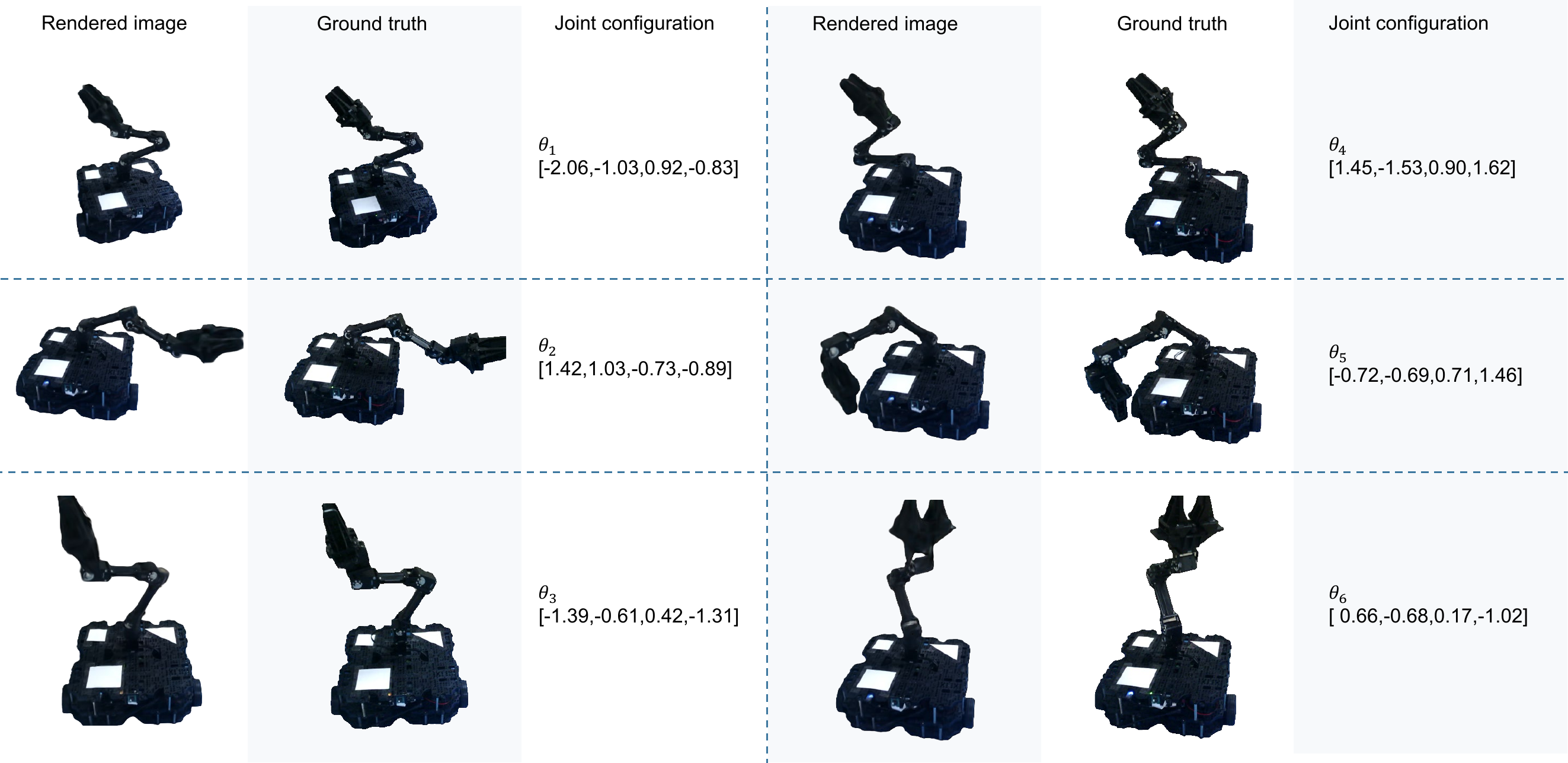}
    \caption{Physical experiments on the OpenManipulator robot with no joint limitation. All joint configurations were randomly generated, and the images were rendered from novel view angles. Joint configurations are shown in radians.}
    \label{fig:real}
\end{figure*}

The results of physical experiments are shown in Figure~\ref{fig:real}. We randomly selected three sets of joint angles that were not in the training data, rendered the corresponding robot images using the trained model, and compared them to the ground truth. The comparison reveals that the proposed method accurately represents the robot's morphology and surface color, further validating the effectiveness of our approach.

\begin{figure}
    \centering
    \includegraphics[width=1\linewidth]{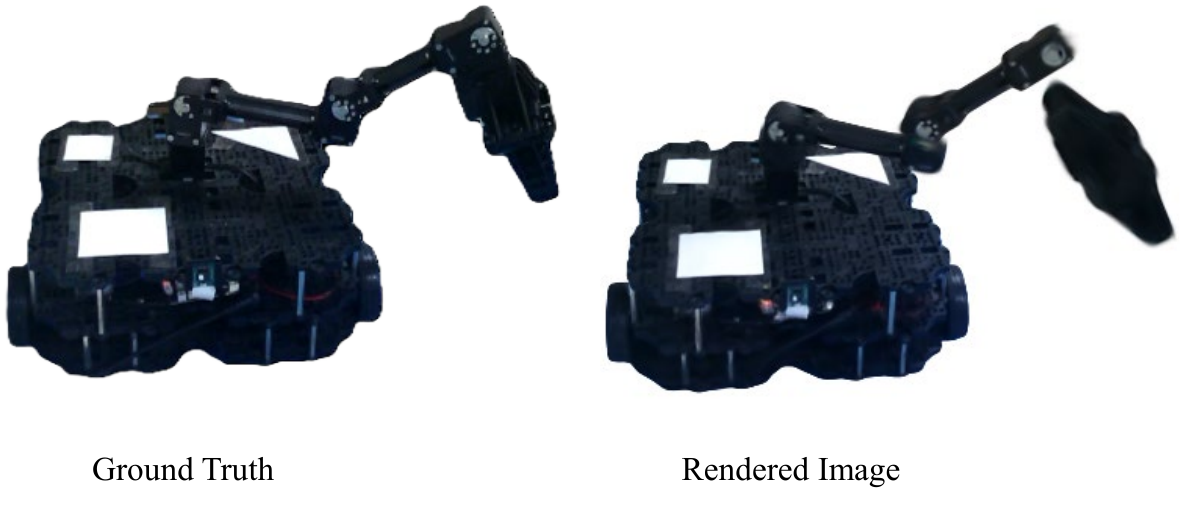}
    \caption{A failure example of the physical experiment.}
    \label{fig:failure}
\end{figure}

Conducting self-modeling with real-world data is more challenging than with simulation data. Thus, the real-world results are not as accurate as those obtained from simulations. This can be attributed to several factors: (1) Potential inaccuracies in the camera's intrinsic and extrinsic parameters calculated by SfM, (2) Imperfect segmentation of the robots from the background in certain images, and (3) A smaller volume of data is available in real-world experiments compared to simulation. A failure example is illustrated in Figure \ref{fig:failure}. In the presented example, the robot's end effector was observed to be detached from its main body. Furthermore, the morphology of the robot's end-effector was also found to be inaccurate. {Possible reasons for this failure include the limited quantity of the images we collected. Another potential reason is our assumption that each link has a similar size, whereas the volumes of the robot's links differ significantly in reality. }

\subsection{Applications}

In this section, we use the proposed self-model to complete some downstream tasks: reaching a target, touching a target with the robot's surface, and reaching a target while avoiding an obstacle. We also show that our self-model can be used to solve the inverse kinematics problem.

\subsubsection{Reaching a target}

A green ball was placed within the robot's workspace, requiring the robot to use its self-model to determine how to make contact with the ball using a specific part of its body. Our neural bone representation simplifies identifying specific parts of the robot and enables targeted control to reach the desired object. We use the robot's end effector as an example. Starting from a resting configuration, the joint angles are progressively adjusted to minimize the distance between the target and the center of the neural bone associated with the end effector. This is accomplished by fixing the self-model parameters and applying gradient descent to automatically optimize the joint angles. In practice, the AdamW~\citep{loshchilov2019decoupledweightdecayregularization} optimizer is utilized, and the loss function is defined as the mean squared error (MSE) between the bone center and the target position:

    \begin{equation}
        L =\frac{1}{3} ||p_b- p_t||_2^2
    \end{equation}

where $p_b$ is the position of the center of the specified bone, and $p_t$ is the position of the target.

In order to keep joint angles within the range limit, we use a sine function to constrain joint values within $[-1, 1]$ {and avoid introducing discontinuities. We use a learning rate of 0.01 to avoid substantial changes in the optimizable joint values.} Then we scale it to the actual range limit. Unlike the method in~\citet{chen2022fully}, by altering the neural bone applied in the loss function, the robot can be controlled to reach the target using different parts without the need for additional training.

\subsubsection{Touching a target with surface}

The above method is only applicable when the target is a point in space rather than an actual sphere, as the target position will end up inside the robot link. Therefore, a different approach is required if the target is an actual sphere. We regard all 3D Gaussians as point clouds. Instead of computing the distance between the center of the target sphere and the center of the neural bone, we compute the distance between the surface of the target sphere and the points belonging to the specific link and minimize it. Again, we use the end effector as an example. We use the skinning weights derived by \eqref{skining weight} of each 3D Gaussian to determine which part they belong to. This process is formulated as

\begin{equation}
    \mathcal{M}_i=\arg\max_{m\in{1, 2, ..., M}}(w_{im}) 
\end{equation}

where $w_i$ is the skinning weights of 3D Gaussian $i$ with respect to all neural bones, and $\mathcal{M}_i$ is the index of the part that 3D Gaussian $i$ belongs to.
Then, the loss function is
\begin{equation}
    L = \min_{\mu \in \{\mu_i|\mathcal{M}_i = \mathcal{M}_{ee}\}}(||p_{target} - \mu||_2) - r
\end{equation}
where $\mathcal{M}_{ee}$ is the index of the end effector, $p_{target}$ is the position of the target and $r$ is radius of the target sphere.
We stop optimization to avoid collision with the target when the minimum distance is less than 0.001.

\begin{figure*}[tbh]
    \centering
    \includegraphics[width=1\linewidth]{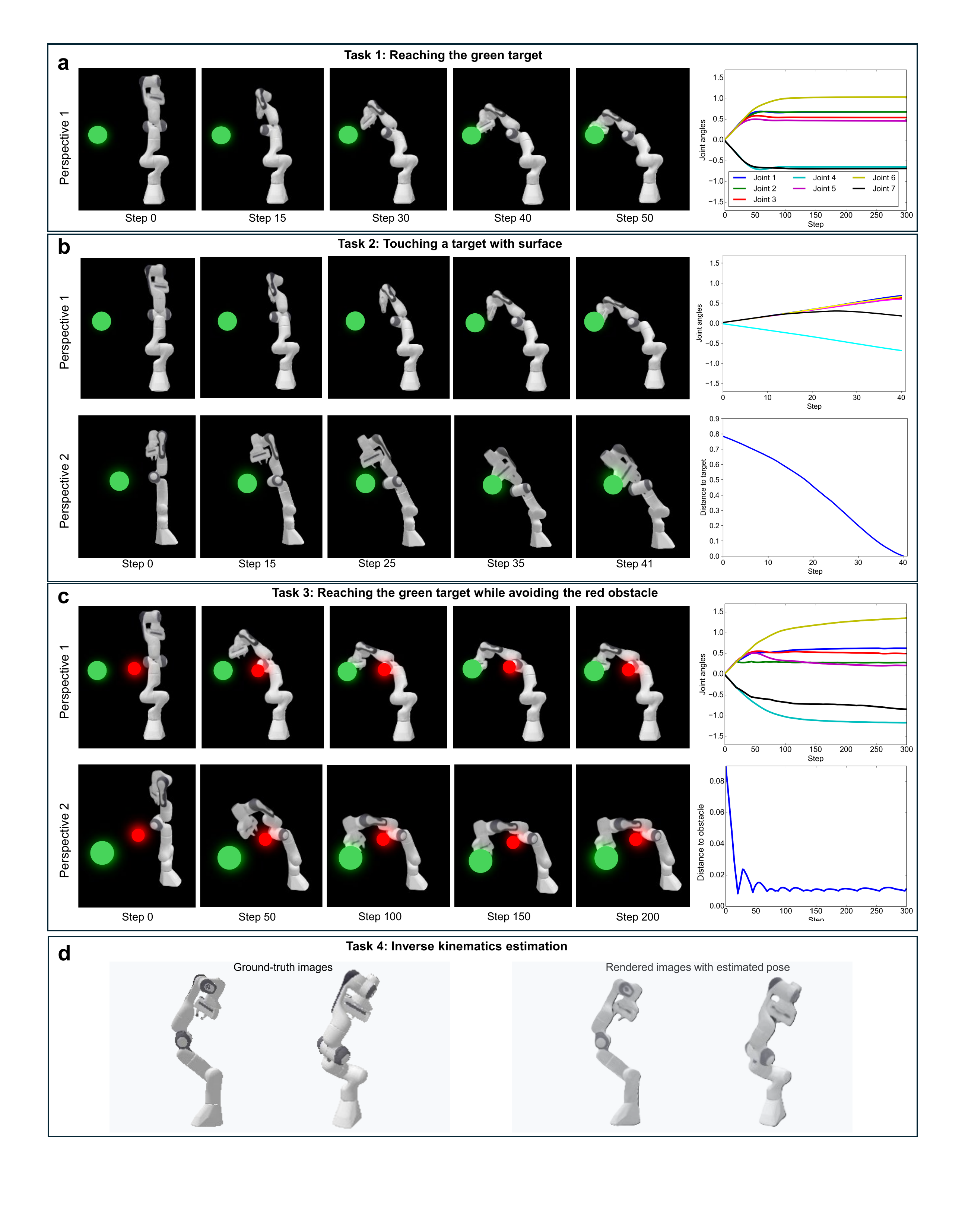}
    \caption{Results of applications. \textbf{a} Snapshots and joint angles of the robot with respect to step number. \textbf{b} Snapshots from different perspectives, joint angles of the robot, and the minimum distance between the target and the end effector of the robot with respect to step number. \textbf{c} Snapshots from different perspectives, joint angles of the robot, and the minimum distance between the obstacle and the robot with respect to step number. The unit of joint angles is radians. At the beginning, all joints of the robot are set to 0. Using solely the self-model, the robot is able to find joint configurations that can fulfill the tasks. \textbf{d} Results of inverse kinematics estimation. The ground truth joint angles are [-0.204, -0.9319, -0.3448, -1.5012, 0.4474, 0.105, 0.3166]. The estimated joint angles are [-0.2127, -0.9321, -0.3051, -1.5050,  0.4863,  0.0258,  0.3113]. Joint angles in radians are slightly different, but the morphology of the robot is close to the ground truth.}
    \label{fig:app1}
\end{figure*}

\subsubsection{Reaching a target while avoiding an obstacle}
An obstacle is placed at a location the robot would pass through after completing the first task of reaching the target. The robot is required to find a new joint configuration to reach the target, as the original configuration would result in a collision with the obstacle. To address this, an additional loss function is introduced alongside the MSE loss function used in the previous task. All 3D Gaussians are treated as point clouds, and the distances $D$ between the points and the obstacle are computed. The loss function is then formulated as
\begin{equation}
    L=\sum_{i=1}^{N}{\max(0, \varepsilon - D_i)}
\end{equation}
where $\varepsilon$ is a threshold that controls how close the robot can get to the obstacle. In our experiment, about 50,000 3D Gaussians were used in the self-model. All 3D Gaussians were forced to keep a distance from the obstacle. {When faced with more challenging situations, such as the presence of multiple obstacles in the space, our proposed simplified method may fail and can not generate a collision-free trajectory. However, since the self-model can compute the distance from each obstacle to each 3D Gaussian ellipsoid, we can modify mature motion planning algorithms to generate reliable, collision-free trajectories in complex situations.}

The above three applications are demonstrated in Figure~\ref{fig:app1}. Images in the first row show how our self-model controls the end effector of the robot to reach the target, shown here as a green ball. The robot's final pose in this task is shown in the fifth image of the first row (Movie S1). In the second and third lines, we illustrate how the robot is controlled to touch the target sphere with its surface (Movie S2 and Movie S3). The joint angles and minimum distance between the 3D Gaussians and the target surface are shown in the last column. Next, an obstacle was placed at the position of the robot's fourth link in the final pose of task 1, and images in the fourth and fifth rows show two perspectives of our self-model guiding the robot to complete the target-reaching task while avoiding the obstacle. Clearly, the robot adopts a new pose to avoid the obstacle and reach the target (Movie S4 and Movie S5). The minimum distance between the obstacle and the robot's 3D Gaussian representations is also shown in the last column of Figure~\ref{fig:app1}(c).

Compared with traditional methods, our method achieves control of unknown robots to accomplish tasks. A traditional method needs to manually create both a kinematic and a geometric model of the unknown robot before the implementation of the control algorithm. However, our method can autonomously execute the complete pipeline from modeling to control.

\subsubsection{Robot inverse kinematics estimation}
With our self-model, we can estimate the robot's inverse kinematics. Given images of the robot, our self-model can predict a joint configuration to pose the robot similarly. After freezing self-model parameters and treating joint angles as optimizable, we employ gradient descent to approximate the joint configuration. The optimization process begins from the rest pose and terminates when the magnitude of change in joint angles falls below a predefined threshold. Comparing rendered and ground truth images, we optimize joint angles to achieve reasonable values. In practice, we use two images to estimate the robot's pose. Our estimation results are illustrated in Figure~\ref{fig:app1}(d). Although estimated joint angles differ slightly from the actual angles, rendered images at the estimated joint angles match ground-truth images. Our self-model effectively poses the robot in alignment with the ground truth. Additional images from varied view angles could further improve joint angle estimation accuracy.

\section{Discussion}

Our method achieves robot self-modeling and can be used to control the robot. However, the method can be further improved in several ways. (1) Our method is based on the assumption of rigid links of robots, which limits its direct applicability to soft and continuous robots. Extending this method to the soft robot is an important future work. (2) A better initialization of our neural bones is possible. The current method requires a predefined number of bones to initialize the structure. This number is an important hyperparameter that affects the training process. We also assume that each bone is similar in size, but in reality, the size of robot links may vary, necessitating multiple bones to represent a single link. (3) Our method may not always accurately reconstruct the robot at certain joint configurations, especially when the joint's movement range is unrestricted. Besides, due to the limitations imposed on joint angles, the range of applications for the models built using our approach is restricted, as some tasks would unavoidably require the robot to move out of the joint constraint. It is worth researching how to obtain a more accurate self-model across all joint configurations. (4) It is important to find a method to map the 3DGS self-model to a mesh model and modify it for compatibility with common robot control pipelines, such as MOVEIT~\citep{coleman2014reducing} or current reinforcement learning toolkits. (5) We need multiple views to reconstruct the model, and the number of views, along with the camera parameters, can affect the final result. Masks of robots are also required. Researching how to self-model the robot with high quality using fewer views and without known camera parameters and masks is a valuable direction to explore. (6) Several other hyperparameters are required to train the self-model, for example, {the weights of the loss components.} Limiting the number of hyperparameters is another task for the future.

\section{Conclusion}

{We propose a 3DGS-based robot self-modeling method. This approach achieves high-quality modeling of the robot's morphology, surface color, and kinematics. Compared to the previous method that requires RGB images only, our method produces results with much better geometric accuracy.} Our method initiates modeling from the robot's rest pose. We employ learnable neural ellipsoid bones of similar volume to represent the underlying articulated structure and kinematics link of the robot. A kinematic network is trained to capture the robot's kinematic properties. Each 3D Gaussian's movement is governed by transformations of neural bones, enabling full model control. Rendering 3D Gaussians into images enables us to visualize the robot's morphology and surface color. Our self-model accurately reconstructs the robot across various joint configurations, achieving high-quality {mesh extractions and} image synthesis.

The self-model developed using our method can be applied in downstream tasks. With this self-model, the robot is capable of solving motion-planning tasks. Our neural bone representation directly models the robot's structure, allowing us to control different parts of the robot by specifying which bone to optimize in the loss function without requiring additional information or training to adjust the self-model. Since our self-model can synthesize high-quality images, it can also estimate the robot's inverse kinematics by {rendering and comparing with ground truth images.}

\begin{dci}
The author(s) declared no potential conflicts of interest with respect to the research, authorship, and/or publication of this article.
\end{dci}

\begin{funding}
The author(s) disclosed receipt of the following financial support for the research, authorship, and/or publication of this article: This work was supported in part by the National Natural Science Foundation of China [grant number 62173352], and in part by the Guangdong Basic and Applied Basic Research Foundation [grant number 2024B1515020104].
\end{funding}

\begin{sm}
    Movie S1: The robot is controlled by its self-model to reach the green target.
    
    Movie S2: The robot is controlled by its self-model to touch the green target sphere with its surface at viewpoint 1.

    Movie S3: The robot is controlled by its self-model to touch the green target sphere with its surface at viewpoint 2.

    Movie S4: The robot is controlled by its self-model to reach the green target while avoiding the red obstacle at viewpoint 1.

    Movie S5: The robot is controlled by its self-model to reach the green target while avoiding the red obstacle at viewpoint 2.
\end{sm}

\bibliographystyle{SageH}
\bibliography{citation}





\end{document}